\pdfoutput=1

\PassOptionsToPackage{table,xcdraw}{xcolor}
\documentclass[11pt]{article}

\usepackage{acl}

\usepackage{times}
\usepackage{latexsym}

\usepackage[T1]{fontenc}

\usepackage[utf8]{inputenc}

\usepackage{microtype}

\usepackage{inconsolata}

\usepackage{amsmath} 
\usepackage{nicefrac}
\usepackage{graphicx}
\usepackage{booktabs}
\usepackage{multirow}
\usepackage{url}
\usepackage{subcaption}
\usepackage[table,xcdraw]{xcolor}

\title{The Curious Decline of Linguistic Diversity: \\Training Language Models on Synthetic Text}

\author{Yanzhu Guo$^1$,  Guokan Shang$^{2,4}$\Thanks{\tiny{Part of the work was done when this author was affiliated with Linagora.}}, Michalis Vazirgiannis$^{1,2}$, Chloé Clavel$^{3}$ \\
$^1$École Polytechnique, Institut Polytechnique de Paris \\ 
$^2$Mohamed bin Zayed University of Artificial Intelligence \\ 
$^3$Inria
$^4$Linagora \\
\texttt{yanzhu.guo@polytechnique.edu}, \texttt{guokan.shang@mbzuai.ac.ae}\\
\texttt{mvazirg@lix.polytechnique.fr},
\texttt{chloe.clavel@inria.fr} }

\begin{document}
\maketitle
\begin{abstract}
This study investigates the consequences of training language models on synthetic data generated by their predecessors, an increasingly prevalent practice given the prominence of powerful generative models. Diverging from the usual emphasis on performance metrics, we focus on the impact of this training methodology on linguistic diversity, especially when conducted recursively over time. To assess this, we adapt and develop a set of novel metrics targeting lexical, syntactic, and semantic diversity, applying them in recursive finetuning experiments across various natural language generation tasks in English. Our findings reveal a consistent decrease in the diversity of the model outputs through successive iterations, especially remarkable for tasks demanding high levels of creativity. This trend underscores the potential risks of training language models on synthetic text, particularly concerning the preservation of linguistic richness. Our study highlights the need for careful consideration of the long-term effects of such training approaches on the linguistic capabilities of language models.
\end{abstract}

\section{Introduction}

\label{sec:intro}

The scaling law reveals a predictable smooth increase in model performance as the amount of data, compute power, and model parameters are increased in tandem \citep{ganguli2022predictability}.
Even assuming that we can boost the other two ingredients indefinitely, the amount of data is limited.
By one estimate, the world’s entire supply of \textit{high-quality} text ranges up to 17 trillion tokens, with a 4-5\% yearly growth rate \citep{villalobos2022will}. This includes all the world’s books, scientific papers, news articles, Wikipedia pages, available code, and the rest of filtered web content. 
Meta’s Llama 2, one of today’s leading LLMs, was trained on around 2 trillion tokens \citep{touvron2023llama}. In other words, we might be approaching the exhaustion of the world’s entire stock of usable language training data, potentially within an order of magnitude.

Is it possible for language models to train on self-generated samples, thereby offering a solution to the looming data shortage?
In fact, whether intentionally or unintentionally, this would happen with the widespread recognition and usage of LLMs.
Regarding pretraining data, which is often sourced from the Internet, a significant trend is occurring: an increasing volume of online content is either generated or assisted by models, and such content is nearly indistinguishable from data produced by humans \citep{uchendu2023does}. Consequently, the subsequent generations of models will inevitably be pretrained on deeply blended data.
Regarding finetuning data, employing LLM-generated examples is already a widely adopted data augmentation approach in the NLP community. The work of self-instruct \cite{wang-etal-2023-self-instruct} prompts language models to solicit synthetic multi-task instruction-tuning data in an iterative bootstrapping way, starting with a seed set of manually-written instructions. Concerning single-task training, \citet{zhou2023multi} build a large-scale dialogue summary corpus annotated by ChatGPT \citep{ouyang2022training} to enhance their pretrained dialogue summarization model.

However, recent studies raise concerns that the above approach of training on \textit{predecessor-generated} text---language models are trained on the synthetic data produced by previous models---is not a panacea without side effects, especially when conducted \textit{recursively} over time. This would introduce a new set of challenging issues, a phenomenon described as \textit{model collapse} \citep{shumailov2023curse,alemohammad2023self}.
On one hand, incorporating model-generated content in training may lead to irreversible flaws in the resulting models, where tails of the original distribution of genuine human content disappear \citep{shumailov2023curse}. On the other hand, even when these models remain free of defects, they could converge to excessively uniform behaviours, with very small variance, due to the recursive sampling of only high probability events. 

In this study, rather than focusing on shifts in task-solving performance, our primary interest lies in exploring changes in language variation caused by the degenerative recursive training process. We target linguistic diversity, a fundamentally important but significantly overlooked aspect of language usage.
Our work is motivated by and contributes to, answering the following two key research questions: First, \textit{how can linguistic diversity be quantified effectively?} Second, \textit{does recursive training on synthetic text result in a reduction of linguistic diversity in model outputs?}

To address these questions, we first develop a comprehensive set of metrics\footnote{Code available at \url{https://github.com/YanzhuGuo/linguistic-diversity}} assessing at three different aspects of linguistic diversity: lexical, semantic, and syntactic. 
Subsequently, we proceed to conduct a series of recursive finetuning experiments spanning three natural language generation tasks, each demanding varying levels of creativity: news summarization \citep{hasan-etal-2021-xl}, scientific abstract generation, and story generation \citep{fan-etal-2018-hierarchical}. 
Our results indicate a notable trend: with the progression of recursive finetuning iterations, there is indeed a remarkable decrease in the diversity of the generated outputs.
This observation highlights the significant impact that training on text generated by predecessors has on the linguistic diversity of language models.

\section{Related Work}

In this section, we explore two avenues of related work: current approaches to evaluate linguistic diversity and recent research on training with synthetic data generated by language models.

\subsection{Evaluating Linguistic Diversity}

Efforts to evaluate language models predominantly concentrate on their performance in task-solving. While some studies extend their scope to include aspects like factual consistency \citep{guo-etal-2022-questioning}, reasoning capability \citep{helwe2021reasoning}, and robustness \citep{chang2023survey}, there is a notable lack of attention paid to linguistic diversity.

Furthermore, the existing studies that do address the diversity issue typically focus on lexical diversity alone. 
For example, in quantifying diversity, research on decoding strategies \citep{li-etal-2023-contrastive, Vijayakumar_Cogswell_Selvaraju_Sun_Lee_Crandall_Batra_2018, ippolito-etal-2019-comparison} usually considers the proportion between the number of unique $n$-grams and total number of $n$-grams in generated text, known as distinct-$n$ metric \citep{li-etal-2016-diversity}.
This very approach can also be found in the literature related to specific NLG tasks, especially those studying creative text generation, such as poetry \citep{chakrabarty-etal-2022-help}, lyric \citep{tian-etal-2023-unsupervised}, and pun \citep{mittal-etal-2022-ambipun} generation. 

Alternatively, \citet{zhang-etal-2021-trading} propose to use Shannon entropy to quantify diversity. However, such an approach is still calculated on tokens (i.e., lexical level), demonstrating a strong correlation with distinct-$n$.
\citet{zhu2018texygen} introduce Self-BLEU which calculates the BLEU similarity score \citep{papineni-etal-2002-bleu} between different sentences of the same document, with higher Self-BLEU implying lower diversity. 
This metric is adopted as a proxy for diversity in evaluating the capability of LLMs in the context of producing content for disinformation operations \citep{liang2022holistic}. 
Nevertheless, the BLEU score is based on $n$-gram overlap and thus also represents diversity solely from the lexical aspect. 

Few works study diversity beyond the lexical level. Recently, \citet{padmakumar2023does} bring up the semantic aspect of diversity and define the average pairwise BERTScore among a set of documents as the homogenization index. They also use ChatGPT to annotate key points on a small set of documents, counting the percentage of unique key points as content diversity. 
\citet{stasaski-hearst-2022-semantic} hypothesize that the semantic diversity can be reflected by the contradictory level---measured by a natural language inference model---among different generation samples given the same input context, while \citet{tevet-berant-2021-evaluating} consider it as the negation of the semantic similarity.
As an exploratory approach to quantify syntactic diversity, \citet{a89efe5d-217a-3260-b2b1-1437ae204234} first manually annotate a small corpus of texts produced by second language learners for syntactic features such as syntactic length and clause types, whose variation is then viewed as a diversity index. 
\citet{huang-etal-2023-paraamr} define the syntactic diversity as the editing distance between the constituency parse trees of two sentences in the context of paraphrase generation.
\citet{mccoy-etal-2023-much} investigate linguistic novelty, in a sense of language generation not simply copies training text, in terms of sequential structure ($n$-grams) and syntactic structure (constituency and dependency).

Our work is the first to comprehensively evaluate text generation on all three aspects of linguistic diversity: lexical, syntactic and semantic, with novel automatic metrics.

\begin{figure*}[ht]
    \centering
    \includegraphics[scale=0.55]{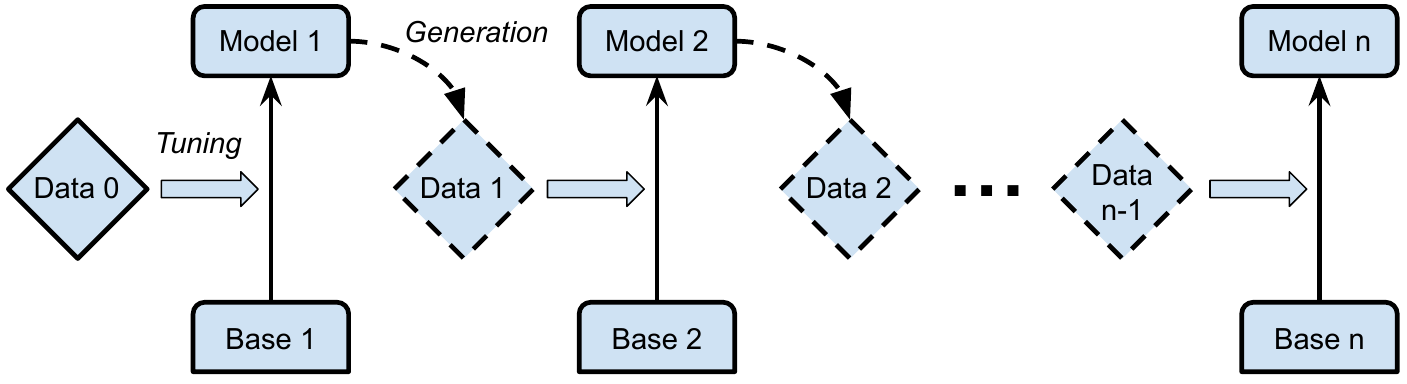}
    \caption{Our recursive tuning-generation process. Beginning with authentic, human-curated Data (0), Base (1) model undergoes finetuning to develop Model (1), which is the first model subject to our language diversity research. Subsequently, we use Model (1) to create synthetic Data (1) to train a successor Model (2) of the next generation, on the basis of Base (2) model. The process continues for n iterations. Base (1), Base (2), ..., Base (n) follow the same model architecture but are independently initialized instances.}
    \label{fig:methodology}
\end{figure*}

\subsection{Training with Synthetic Text}

Ever since the introduction of generative adversarial networks \citep{NIPS2014_5ca3e9b1}, training new models with synthetic data produced by various generators has become a means of data augmentation \citep{LI202271}, a practice that has been expanding to all modalities of machine learning research, including image, audio, and text.

However, the large-scale usage of this approach, particularly employing tremendous quantities of synthetic text to train generative models, is a more recent trend \citep{dai2023chataug,marwala2023use}.
To name a few, the self-instruct study by \citet{wang-etal-2023-self-instruct} guides a language model to iteratively generate synthetic multi-task instruct-tuning data, beginning with an initial set of manually-written instructions.
\citet{huang2022large} demonstrate that LLMs are capable of self-improving their reasoning abilities, with generated high-quality answers for unlabeled questions, using chain-of-thought \citep{wei2022chain} and self-consistency \citep{wang2022self} prompting techniques.
Meanwhile, \citet{xu2023baize} introduce a pipeline that autonomously generates a high-quality, multi-turn chat corpus by enabling ChatGPT to converse with itself, which is then used to enhance a LLaMA model. 

As already mentioned in Section \ref{sec:intro}, 
studies show that this training methodology will eventually lead to model collapse \citep{shumailov2023curse} when conducted recursively, causing performance degeneration, regardless of potential data filtering or refinement \citep{alemohammad2023self}.
Our research is motivated by the same concept, but we focus on investigating the impact of recursive training on linguistic diversity instead of performance. To the best of our knowledge, our work is the first to address this issue.

\section{Methodology}

This section introduces our recursive training methodology and outlines the  linguistic diversity metrics. 

\subsection{Recursive Training Simulation}\label{sec:simulation}
Following the work of \citet{shumailov2023curse}, we simulate the process of recursively training language models on predecessor-generated text, under a finetuning setting.
As illustrated in Figure \ref{fig:methodology}, we begin with human-generated task-finetuning Data (0), which is used to train Base (1) model to create a task-specialized version, referred to as Model (1).
After that, we use Model (1) to produce synthetic task-finetuning Data (1), which serves to train the next generation, Model (2), built upon Base (2) model.
This procedure is repeated $n$ times.

For the sake of simplicity, we start from a new instance of the same base model across different generations, i.e., Base (1) = Base (2) = $,...,$ = Base ($n$).
In addition, we only use Data ($n-1$) to train Model ($n$), whereas in a setting closer to the real-life scenario, we have access to the accumulated data ensemble of all predecessors, i.e., Data $\{$(0)$,$ (1) $,...,$ ($n-1$)$\}$. This simplification draws from the results of \citet{shumailov2023curse}, which indicates that model collapse is unavoidable, even when the training involves the full ensemble of accumulated data, though the effect is somewhat attenuated.

In terms of finetuning tasks, we chose three distinct natural language generation tasks, each characterized by varying degrees of constraint, from the most restrictive to the least: news summarization, where summaries must closely align with the original content; scientific abstract generation, with some initial context provided, but room for creative expansion; and story generation, which allows for the most creativity and freedom in expression.

In the end, we conduct our linguistic diversity research with the finetuned Model $\{$(1)$,$ (2) $,...,$ ($n$)$\}$ for each task, subjecting them to evaluation on the test set of the corresponding task.

\subsection{Perplexity}
Our research primarily focuses on linguistic diversity, yet we also require a reliable metric to verify that our finetuned models are well-aligned with the training data. Perplexity, a standard metric for assessing language modeling, evaluates a model's level of ``surprise'' or ``confusion'' when encountering a given sequence of tokens. Models that more accurately mirror the training data's distribution exhibit lower perplexity. While useful for model comparison, perplexity doesn't fully reflect text quality \citep{meister-etal-2023-locally}. A low perplexity score suggests higher predictive precision, but texts can be grammatically sound and contextually coherent yet still score high in perplexity if they include unusual or creative language not included in the model's training data \citep{basu2021mirostat}. In our study, a model with lower perplexity is not deemed superior by default. Our aim is to ensure that the perplexity remains within a reasonable limit, producing texts of sufficient quality for our linguistic diversity evaluation.

\subsection{Linguistic Diversity Metrics}

We approach the evaluation of linguistic diversity from three different perspectives: lexical diversity, semantic diversity and syntactic diversity.

\subsubsection{Lexical Diversity}

Lexical diversity metrics are used to measure the variety of words used in a text, which is contended to mirror the extent of vocabulary possessed by a writer or speaker. We believe a degenerated language model, which presumably has a smaller vocabulary, will use a narrower variety of lexical items than non-degenerated language models. We select different metrics operating at different levels of textual granularity: word, $n$-gram, and sentence. 

\medskip

\noindent \textbf{Type-Token Ratio (TTR)} \citep{johnson1944studies,templin1957certain}, the most well-known metric, which is calculated as the number of unique words (types) $t$ divided by the number of running words (tokens) $c$, i.e., $\text{TTR} = \nicefrac{t}{c}$. This metric was used to study the language development in child language research, a low value is probably indicative of a language-specific deficiency \citep{miller1981assessing}. The length of a text inherently skews vanilla TTR values, with longer texts generally yielding lower TTR scores due to an inexorably decreased occurrence of unique novel words (drawn from a limited vocabulary) as the text lengthens \citep{richards1987type}. Following common practice \citep{shaib2024standardizing}, we truncate all texts to a fixed length before computing the TTR \footnote{Details on the truncation lengths can be found in Appendix B.}.

\medskip

\noindent\textbf{Distinct-}$\mathbf{n}$ \citep{li-etal-2016-diversity}, which equals the proportion between the number of unique $n$-grams and total number of $n$-grams in tested text \citep{xing2017topic}.
This metric is originally introduced in the context of enhancing the response diversity of conversational agents, which frequently produce safe and fluent but dull and uninformative generic responses at time (e.g., \textit{I don't know}) \citep{han-etal-2022-measuring}. Similar to naive TTR, distinct-$n$ varies as a function of text length, so we report the results at fixed sizes for $n=2$ and $n=3$ (distinct-$1$ is equivalent to TTR). 


\medskip

\noindent\textbf{Self-BLEU} \citep{zhu2018texygen}, a recently developed method for evaluating the diversity of synthetic text. This method assesses the similarity between one sentence and the rest in a group of generated sentences. It treats one sentence as the hypothesis and the others as references to calculate BLEU scores \citep{papineni-etal-2002-bleu}. The final Self-BLEU score averages these BLEU scores across all generated sentences. We report $1-\texttt{Self-BLEU}$, so a higher value reflects richer diversity of the generation \citep{palumbo-etal-2020-semantic}.

\subsubsection{Semantic Diversity}

According to recent studies \citep{tevet-berant-2021-evaluating, stasaski-hearst-2022-semantic}, the above lexical-level metrics often fail to capture semantic diversity, since texts including similar words can have different semantics and texts with different words can have similar semantics \citep{pmlr-v80-yarats18a}. We tackle this problem by transforming sentences into semantically meaningful sentence embeddings using Sentence-BERT \cite{reimers-gurevych-2019-sentence}. We quantify semantic diversity as the dispersion of sentence embeddings over the semantic space. The dispersion is measured by the average pairwise cosine-distance of all embedding vectors (\texttt{Div\_sem}).

\subsubsection{Syntactic Diversity}

The significance of syntactic diversity is often underestimated in NLP, despite its importance. For language learners (as well as language models), exposure to a wide range of syntactic structures is beneficial for developing a more comprehensive understanding of the language \citep{aggarwal-etal-2022-towards}. Moreover, a range of syntactic forms enhances expressiveness and subtlety in writing, influencing the style and tone of a text \citep{edwards1998diversity}. While linguistic and language acquisition research \citep{a89efe5d-217a-3260-b2b1-1437ae204234} has explored this aspect, these studies typically rely on manual annotation of features, a process that can be costly and prone to human error.

We introduce the first graph-based metric to quantify syntactic diversity. We use a neural parser \citep{qi-etal-2020-stanza} to construct dependency trees from  sentences, following the universal dependencies formalism. These trees are then transformed into graph representations, with nodes representing the words and edges indicating the dependency relationships between them. Subsequently, we employ the Weisfeiler-Lehman graph kernel \citep{10.5555/1953048.2078187, JMLR:v21:18-370} to map these graphs into a vector space. This kernel, rooted in the Weisfeiler-Lehman isomorphism test, effectively positions graphs that are structurally alike closer to each other in the embedding space. To assess syntactic diversity, we calculate it similarly to semantic diversity, using the average pairwise distance (\texttt{Div\_syn}).

\begin{table*}[!ht]
\footnotesize
\begin{center}
\begin{tabular}{ |p{0.97\linewidth}| }
\hline 
\textcolor{violet}{\textbf{Prompt}: } 
The title of the paper is: {CASIA}{'}s System for {IWSLT} 2020 Open Domain Translation.The abstract of the paper is: This paper describes the CASIA{'}s system for the IWSLT 2020 open domain translation task.
\smallskip \\
\hline
\textcolor{purple}{\textbf{Human}: }
This year we participate in both Chinese→Japanese and Japanese→Chinese translation tasks. Our system is neural machine translation system based on Transformer model. We augment the training data with knowledge distillation and back translation to improve the translation performance. Domain data classification and weighted domain model ensemble are introduced to generate the final translation result. We compare and analyze the performance on development data with different model settings and different data processing techniques..
\smallskip \\
\hline
\textcolor{purple}{\textbf{Iteration 1}: }
 We participated in all language directions, namely Chinese $\leftrightarrow$ English, Japanese$\leftrightarrow$ English and German $\leftrightarrow$ English. Our system mainly focuses on data selection, data preprocessing and data filtering to improve the quality of the generated translations.. 
\smallskip \\
\hline
\textcolor{purple}{\textbf{Iteration 3}: }
We participated in all three language pairs, namely Chinese$\leftrightarrow$English, Japanese$\rightarrow$Chinese and German$\leftrightarrow$French. Our systems are based on Transformer architecture with a simple but effective method to improve the translation quality by reusing monolingual data from other languages.. The source code can be obtained from \url{https://github.com/csebuetnlp/acl2021/opendomaintranslate}..The blind test set will be released upon the acceptance of the report..The full results can also be found at \url{https://competitions.codalab.org/competitions/26153}.. The leaderboard can be found at \url{https://competitions.codalab.org/page/view/11650}..The dataset used in this work has been released under CC BY 4.0 license..The experimental results show that our proposed methods significantly outperform the baseline models..
\smallskip \\
\hline
\textcolor{purple}{\textbf{Iteration 5}: }
We participated in all three language pairs, namely Chinese$\leftrightarrow$English (Chinese$\rightarrow$German), Japanese$\rightarrow$Czech and German$\leftrightarrow$French.. Our approach relies on a Transformer model trained with in-domain data only; it also incorporates a novel method to improve the quality by leveraging parallel corpora from other domains such as TED Talks. The code used in this work has been released under CC BY 4.0 license at \url{https://github.com/czyssrs/ibm_nlp2020/}. For further information about our systems please see \url{http://www.csse.unige.edu/research/demo/}.. The full text of this report can be obtained via \url{https://competitions.codalab.org/page/view/11153}.. The codes for our experiments are available online at \url{https://codebenchmark.io/}, too....... Finally, we provide some insights into how our systems perform when translating into low resource conditions...... The source code for our systems is publicly available through \url{https://github.com/czeneva/acl2022tasks2/}. \\

\hline
\end{tabular}
 \caption{
An example of model outputs over iterations. Since diversity is a holistic concept observed among the entire set of documents, it is non evident to demonstrate the loss of diversity through a single example. However, for the scientific abstract generation task in this example, there are some obvious patterns. Abstracts found in the ACL anthology often contain URLs linking to data, code or demos. After a few iterations of training on synthetic text, the model picked up this high probability event and starts to generate such links repeatedly in an unnatural manner.}
\label{example}
\end{center}
\end{table*}

\section{Experiments and Results}

We conduct our experiments on three generative tasks, as introduced in Section \ref{sec:simulation}, with decreasing degrees of constraint and increasing degrees of creativity: abstractive news summarization, scientific abstract generation, and story generation. 

\subsection{Experimental Setup}
For each task, we simulate 6 iterations of the recursive training chain, i.e., $n=6$ in Figure \ref{fig:methodology}. 
Following previous work \cite{shumailov2023curse}, we select OPT \citep{zhang2022opt} as our base model, and each iteration begins with a new instance of the base model. Different from \citet{shumailov2023curse}, we use OPT-350M instead of OPT-125M to maintain higher generation quality over iterations, avoiding excessive noise.
Model (1) is finetuned on Data (0)---the training set of the finetuning task---which is human-authored. 
From Model (2) to Model (6), they are finetuned on synthetic Data ($n-1$) generated by their predecessor Model ($n-1$).
We go through all of the original training examples in Data (0) to produce a comparable synthetic dataset of the same number of samples.
The models are finetuned for 5 epochs using the AdamW optimizer \citep{loshchilov2017decoupled} on a cluster of two NVIDIA RTX A6000 GPUs. 

In the following, we explain each of the three tasks in detail.


\begin{table*}[ht]
\centering
\resizebox{0.8\textwidth}{!}{%
\begin{tabular}{@{}llccccccc@{}}

\toprule
 &
  \textbf{Iter} &
  \textbf{PPL} &
  {\color[HTML]{3531FF} \textbf{TTR}} &
  {\color[HTML]{3531FF} \textbf{Distinct-2}} &
  {\color[HTML]{3531FF} \textbf{Distinct-3}} &
  {\color[HTML]{3531FF} \textbf{1-Self-BLEU}} &
  {\color[HTML]{8612E7} \textbf{Div\_syn}} &
  {\color[HTML]{BB24CC} \textbf{Div\_sem}} \\ \midrule
 &
  {\color[HTML]{680100} Human} &
  {\color[HTML]{680100} --} &
  {\color[HTML]{680100} 7.36} &
  {\color[HTML]{680100} 48.1} &
  {\color[HTML]{680100} 81.1} &
  {\color[HTML]{680100} 73.3} &
  {\color[HTML]{680100} 3.17} &
  {\color[HTML]{680100} 46.6} \\
 &
  1 &
  12.5 &
  5.99 ($\downarrow$)&
  37.9 ($\downarrow$)&
  68.5 ($\downarrow$)&
  74.6 ($\uparrow$)&
  1.65 ($\downarrow$)&
  47.2 ($\uparrow$)\\
 &
  2 &
  3.42 &
  5.55 ($\downarrow$)&
  35.5 ($\downarrow$)&
  64.1 ($\downarrow$)&
  74.2 ($\downarrow$)&
  1.76 ($\uparrow$)&
  47.2 ($\rightarrow$)\\
 &
  3 &
  3.09 &
  4.99 ($\downarrow$)&
  32.6 ($\downarrow$)&
  59.3 ($\downarrow$)&
  72.6 ($\downarrow$)&
  1.95 ($\uparrow$)&
  46.8 ($\downarrow$)\\
 &
  4 &
  2.86 &
  4.46 ($\downarrow$)&
  29.2 ($\downarrow$)&
  54.5 ($\downarrow$)&
  69.7 ($\downarrow$)&
  1.85 ($\downarrow$)&
  46.6 ($\downarrow$)\\
 &
  5 &
  2.62 &
  3.92 ($\downarrow$)&
  25.8 ($\downarrow$)&
  49.5 ($\downarrow$)&
  68.0 ($\downarrow$)&
  1.62 ($\downarrow$)&
  46.0 ($\downarrow$)\\
\multirow{-7}{*}{\textbf{\begin{tabular}[c]{@{}l@{}}News \\ Summarization\end{tabular}}} &
  6 &
  2.48 &
  3.66 ($\downarrow$)&
  25.6 ($\downarrow$)&
  49.2 ($\downarrow$)&
  65.3 ($\downarrow$)&
  0.82 ($\downarrow$)&
  46.6 ($\uparrow$)\\ \midrule
 &
  {\color[HTML]{680100} Human} &
  {\color[HTML]{680100} --} &
  {\color[HTML]{680100} 3.09} &
  {\color[HTML]{680100} 35.4} &
  {\color[HTML]{680100} 75.0} &
  {\color[HTML]{680100} 71.0} &
  {\color[HTML]{680100} 4.52} &
  {\color[HTML]{680100} 40.4} \\
 &
  1 &
  13.4 &
  2.06 ($\downarrow$)&
  20.7 ($\downarrow$)&
  48.3 ($\downarrow$)&
  64.2 ($\downarrow$)&
  3.80 ($\downarrow$)&
  39.4 ($\downarrow$)\\
 &
  2 &
  3.87 &
  1.96 ($\downarrow$)&
  17.4 ($\downarrow$)&
  39.8 ($\downarrow$)&
  60.4 ($\downarrow$)&
  4.06 ($\uparrow$)&
  38.6 ($\downarrow$)\\
 &
  3 &
  2.59 &
  1.90 ($\downarrow$)&
  16.1 ($\downarrow$)&
  36.0 ($\downarrow$)&
  59.2 ($\downarrow$)&
  4.94 ($\uparrow$)&
  38.6 ($\rightarrow$)\\
 &
  4 &
  2.31 &
  1.82 ($\downarrow$)&
  15.3 ($\downarrow$)&
  34.0 ($\downarrow$)&
  58.7 ($\downarrow$)&
  4.60 ($\downarrow$)&
  37.6 ($\downarrow$)\\
 &
  5 &
  2.24 &
  1.77 ($\downarrow$)&
  14.2 ($\downarrow$)&
  31.6 ($\downarrow$)&
  58.2 ($\downarrow$)&
  4.41 ($\downarrow$)&
  37.5 ($\downarrow$)\\
\multirow{-7}{*}{\textbf{\begin{tabular}[c]{@{}l@{}}Scientific \\ Abstract \\ Generation\end{tabular}}} &
  6 &
  2.17 &
  1.69 ($\downarrow$)&
  13.3 ($\downarrow$)&
  29.5 ($\downarrow$)&
  57.5 ($\downarrow$)&
  4.10 ($\downarrow$)&
  37.1 ($\downarrow$)\\ \midrule
 &
  {\color[HTML]{680100} Human} &
  {\color[HTML]{680100} --} &
  {\color[HTML]{680100} 2.23} &
  {\color[HTML]{680100} 30.5} &
  {\color[HTML]{680100} 70.6} &
  {\color[HTML]{680100} 67.0} &
  {\color[HTML]{680100} 4.84} &
  {\color[HTML]{680100} 43.7} \\
 &
  1 &
  14.1 &
  0.84 ($\downarrow$)&
  13.8 ($\downarrow$)&
  44.2 ($\downarrow$)&
  61.6 ($\downarrow$)&
  4.23 ($\downarrow$)&
  41.4 ($\downarrow$)\\
 &
  2 &
  4.41 &
  0.72 ($\downarrow$)&
  13.3 ($\downarrow$)&
  43.1 ($\downarrow$)&
  61.0 ($\downarrow$)&
  3.41 ($\downarrow$)&
  42.5 ($\uparrow$)\\
 &
  3 &
  3.37 &
  0.68 ($\downarrow$)&
  12.8 ($\downarrow$)&
  42.0 ($\downarrow$)&
  60.6 ($\downarrow$)&
  2.99 ($\downarrow$)&
  43.3 ($\uparrow$)\\
 &
  4 &
  2.99 &
  0.65 ($\downarrow$)&
  12.3 ($\downarrow$)&
  40.9 ($\downarrow$)&
  60.5 ($\downarrow$)&
  2.50 ($\downarrow$)&
  43.3 ($\rightarrow$)\\
 &
  5 &
  2.82 &
  0.63 ($\downarrow$)&
  11.8 ($\downarrow$)&
  39.7 ($\downarrow$)&
  60.5 ($\rightarrow$)&
  2.14 ($\downarrow$)&
  42.7 ($\downarrow$)\\
\multirow{-7}{*}{\textbf{\begin{tabular}[c]{@{}l@{}}Story \\ Generation\end{tabular}}} &
  6 &
  2.70 &
  0.61 ($\downarrow$)&
  11.4 ($\downarrow$)&
  38.6 ($\downarrow$)&
  60.3 ($\downarrow$)&
  1.96 ($\downarrow$)&
  42.5 ($\downarrow$)\\ \bottomrule

\end{tabular}%
}

\caption{Perplexity (PPL) and linguistic diversity metrics for texts generated over different iterations (iter). All diversity metrics range from 0 to 1 and are reported as percentages (cosine distances are halved to maintain this range). Typical values are suggested by the results obtained on human written texts. The arrows in parentheses indicate the direction of variation compared to the previous iteration. In the case of iteration 1, the comparison is against human reference text.
}
\label{fig:Results}

\end{table*}

\subsection*{Task1: Abstractive News Summarization}
For abstractive news summarization, we use the XL-SUM \citep{hasan-etal-2021-xl}, one of the most recently proposed datasets. In comparison to the other prominent news summarization datasets, XL-SUM is more abstractive than CNN/DailyMail \citep{NIPS2015_afdec700, cheng-lapata-2016-neural} and more factual than XSUM \citep{narayan-etal-2018-dont, guo-etal-2022-questioning}. It is also larger in scale, consisting of 306,522 samples in the training set and 11,535 samples in the test set. The average length of the news articles is 386 tokens, while the average length of the summaries is 28 tokens. This is a generation task with ``low entropy'' since there is abundant context and the content is restricted.

\subsection*{Task2: Scientific Abstract Generation}
For scientific abstract generation, we parse the bibliography database (BibTeX) file with abstracts of ACL Anthology\footnote{\url{https://aclanthology.org}} in June, 2023. ACL Anthology hosts papers published at computational linguistics or natural language processing venues since 1965. We split the bibliography entries into the training and the test set, resulting in 40,512 samples for train and 2,132 samples for test. We use the title of the paper and the first sentence of the abstract as the prompt, asking the model to finish generating the rest of the abstract. The prompt (title + first sentence) is 42 tokens long on average, while the mean of the full abstract length is 145 tokens. This is a task of ``medium entropy'', the provided title already lays out the general idea of the paper and the first sentence provides a fair amount of context.

\subsection*{Task3: Story Generation}
For story generation, we use the WritingPrompts dataset \citep{fan-etal-2018-hierarchical}. It is made up of human written stories paired with writing prompts from Reddit’s WritingPrompts forum. There are 272,600 samples in the training set and 15,138 samples in the test set. The writing prompts consist of 30 tokens on average and the resulting stories have a mean of 389 tokens. The prompts are generally short and in most cases do not contain a plot (i.e. narrative structure), making this a ``high-entropy'' generation task with limited context.

\subsection*{Decoding Strategy.}
We use a combination of nucleus sampling ($p$) and temperature sampling ($\tau$) to achieve nuanced control over the language model's outputs \citep{Holtzman2020The}. Nucleus sampling, also known as top-$p$ sampling, is used to generate text by selecting the most probable words from a distribution of words. It ensures that the cumulative probability of the chosen words exceeds a certain threshold ($p$). Higher values of $p$ lead to more deterministic text. Temperature sampling involves dividing the output logits by a temperature parameter ($\tau$) before sampling from the distribution. Higher values of $\tau$ make the distribution more uniform and increase randomness.

We adapt the specific parameters to the characteristic of each task \citep{amini-etal-2023-generating}. For news summarization, we emphasize precision and set $p=0.1$, $\tau=0.3$. For story generation, we care more about creativity and set $p=0.9$, $\tau=0.7$. For scientific abstract generation, we search something in between and set $p=0.5$, $\tau=0.5$. The \texttt{max\_new\_tokens} value is chosen according to the length of human-written references for each task: 50 for news summarization, 500 for story generation and 300 for scientific abstract generation. While the decoding strategy has influence over diversity metrics, it is not the determinant factor \cite{giulianelli-etal-2023-comes}. We aim to draw generalizable conclusions by experimenting across three distinct sets of decoding strategies.

\subsection{Results}
\label{results}

In Table \ref{fig:Results}, we display the perplexity and linguistic diversity metrics for texts generated across various iterations. We also show an example of generated texts across iterations in Table \ref{example}.

Our findings indicate that the \textit{perplexity} values fall within an acceptable range ($<$ 20) \citep{Holtzman2020The}, indicating that models effectively assimilate training data and generate texts of a quality viable for diversity analysis. The decrease of perplexity over iterations suggest that the model might be more prone to over-fitting when trained on synthetic text, losing the tail of the original distribution (visualized in Appendix \ref{sec:appendix1}, Figure \ref{fig:tail}). A general decline in the majority of \textit{linguistic diversity} metrics underscores the pressing issue of diminishing linguistic diversity. We highlight some key observations in below.

\paragraph{The decline of diversity is greater for ``high entropy'' tasks.}
We deliberately select three generation tasks of varying ``entropy'', which is reflected by the amount and nature of given context, i.e., constraint. News summarization involves a lengthy context with the summary confined to a very limited space, whereas story generation is characterized by brief prompts and a vast array of potential narrative directions. In the ``highest entropy'' task, story generation, the gap in linguistic diversity between human-written and model-generated texts is the most pronounced, and the decline over iterations is the fastest. The significant gap between humans and models is expected, given that story generation demands substantial creativity, a domain where language models are known to fall short \citep{chakrabarty2023art}. The rapid decrease in diversity can also be explained by the creative nature of the task. Models initially learn from diverse original human-written stories but suffer greatly when later exposed solely to synthetic data, which already exhibits a notable loss in diversity.

\begin{figure*}[ht!]
    \centering
    \begin{subfigure}{.45\textwidth}
        \centering
        \includegraphics[width=\linewidth]{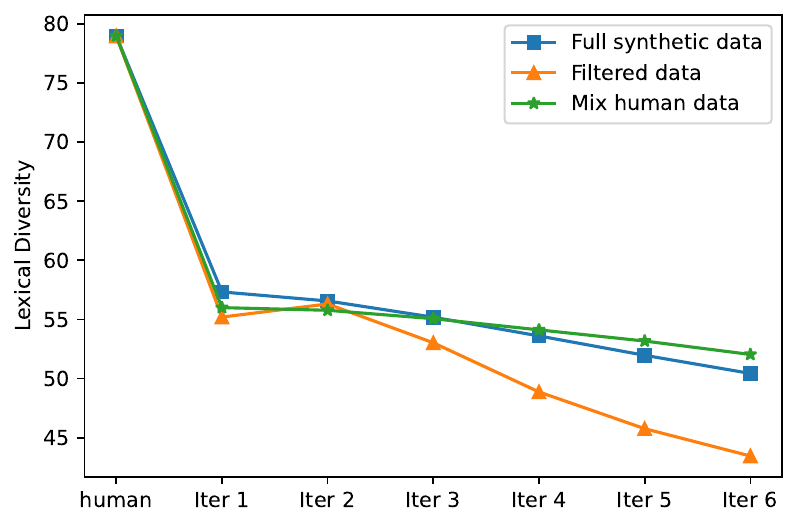}
        \caption{Lexical diversity.}
    \end{subfigure}%
    \hfill
    \begin{subfigure}{.45\textwidth}
        \centering
        \includegraphics[scale=0.535]{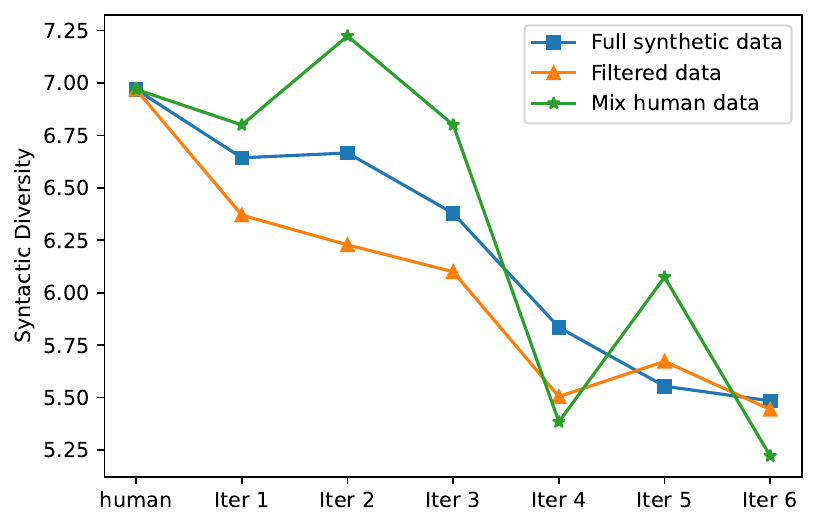}
        \caption{Syntactic diversity.}
    \end{subfigure}

    \caption{Illustration of linguistic diversity variation for the story generation task under different recursion settings. Since there is a strong correlation between different diversity metrics of the same aspect, we only report one per aspect: Distinct-3 for lexical diversity and D\_syn\_c for syntactic diversity.}

    \label{fig:analysis}
    
\end{figure*}

\paragraph{Even for ``lower entropy'' tasks, training on synthetic texts will eventually lead to vanishing diversity.}
In tasks like news summarization and scientific abstract generation, which have ``lower entropy'' compared to story generation, there is still a noticeable decrease in linguistic diversity over iterations. Consider the task of generating scientific abstracts: initially, the syntactic diversity in texts created by Models (2) and (3) shows an increase compared to those written by humans. This might be because scientific abstracts inherently possess less varied syntactic structures than the broader range of texts in the pretraining data of OPT-350M. However, as the iterations advance, the syntactic diversity scores of the texts produced by the models eventually decline, dropping below those of human-written abstracts. This trend might be partly attributed to catastrophic forgetting \cite{MCCLOSKEY1989109}. Additionally, while human-written abstracts may have limited syntactic diversity, their structure is markedly different from the pretraining data, thus introducing new learning elements for the model. In contrast, the synthetic data produced by Model (2), despite its marginally higher internal syntactic diversity, closely mirrors the model's own training distribution. This lack of novel information leads to a subsequent reduction in variation.

\paragraph{Syntactic diversity suffers remarkably.}

We notice that syntactic diversity manifests a decreasing trend, especially for creative tasks, comparable to the decline in lexical diversity and to a greater extent than in semantic diversity (also visualized in Appendix \ref{sec:appendix1}, Figure \ref{fig:syn} and Figure \ref{fig:sem}). While the reduction in lexical diversity is well-researched and somewhat anticipated, our study is the first to highlight the decrease in syntactic diversity. Syntax is more implicit but equally important as vocabulary in maintaining linguistic richness. The important yet overlooked decline in syntactic diversity emphasizes the need for future NLG research to include syntactic diversity measurements alongside the commonly reported lexical diversity metrics.

\paragraph{Semantic diversity is the most stable.}
Semantic diversity remains more stable compared to lexical diversity and syntactic diversity. We believe that synthetic training data have more impact on the diversity of form than the diversity of content. The main issue in the semantic aspect is coherence rather than diversity. We find that the generated texts remain rather diverse in meaning throughout the iterations whereas the coherence between sentences drops. This finding also corresponds to the well-known fact that language models are prone to hallucination \citep{huang2023survey}.

\subsection{Playing with Recursion Settings}
We perform further analysis to understand how different factors influence outputs of the recursive tuning-generation process. We introduce two settings to approximate the real-life scenario. We focus our analysis on the story generation task as it shows the most pronounced diversity decline.

\paragraph{Filtering Synthetic Data.}
Instead of using the full set of synthetic samples, it is a common choice to filter out invalid samples before training \cite{wang-etal-2023-self-instruct}. In our case, we use a \textit{linguistic acceptability} filter to discard the noisy samples generated in each iteration. The filter is a RoBERTa model \footnote{textattack/roberta-base-CoLA} \citep{morris-etal-2020-textattack} trained on the COLA corpus \citep{warstadt-etal-2019-neural}. We do generation on the full training set and discard the 20\% of synthetic data with the lowest linguistic acceptability score before using them to train the next iteration's model.

\paragraph{Mixing Fresh Human Data.}
To approximate the most realistic scenario, we consider mixing human data with synthetic data for training. We separate the training data into a 40\% set reserved for synthetic data generation and a 60\% set used only as human data. The 60\% set of human data is further split into six subsets each containing 10\% of the human data. For each of the 6 training iterations, one of these six 10\% subsets are mixed into the synthetic data as ``fresh human data''. These data are considered ``fresh'' because they were held out at the beginning and not seen by models from previous iterations.

The results are displayed in Figure \ref{fig:analysis}. We do not show semantic diversity as it remains relatively stable across all settings. The introduction of fresh human data only minimally mitigates the observed decrease, while filtering significantly amplifies the decline. This outcome is unsurprising, as quality filters typically favor more common and less inventive samples. Consequently, in practice, the linguistic diversity decline might be even more substantial than suggested by our previous findings.

\section{Conclusion}
Our study provides critical insights into the implications of recursively training language models on synthetic data generated by their predecessors.
Through our innovative approach, focusing on linguistic diversity measures rather than traditional performance metrics, across various NLG tasks, we have uncovered a noticeable reduction in lexical and syntactic diversity in language model outputs over successive iterations of recursive training on synthetic text. These findings highlight a concerning trend: as language models increasingly rely on predecessor-generated text for training, there is a tangible risk of diminishing linguistic richness and variety in their outputs. Our research underscores the necessity for a more nuanced and forward-thinking approach in the development of language models, emphasizing the importance of preserving linguistic diversity alongside improving technical performance.


\section*{Limitations}
\paragraph{Language diversity.}
Our work investigates linguistic diversity in a monolingual context. Our experiments are exclusively conducted in the English language. While the main research idea is readily adaptable, the specific methodologies require adjustments when applied to other languages. It's worth noting that our linguistic diversity metrics may not perform optimally for languages apart from English. These metrics rely on language-specific tokenization/segmentation, dependency parsing, and sentence embeddings, which pose challenges for languages with limited resources. However, it would be interesting future work to overcome these obstacles and investigate linguistic diversity in a multilingual setting.

\paragraph{Resource constraint.}
Due to resource limitations, we could not perform experiments on an extensive range of models. We opted for the moderately large decoder-only model OPT-350M, striking a balance between generation quality and parameter scale. Our analysis involves recursive model training across six iterations, for three tasks and under various settings, demanding significant computational resources. For instance, completing all six iterations for the story generation task under the full synthetic setting alone consumes approximately 700 GPU hours on the NVIDIA RTX A6000 48G GPU. In this study, our primary focus is on comparing different tasks and settings rather than across various models. Nevertheless, we anticipate that the decline in linguistic diversity is a recurring phenomenon in different language models. In future research, we intend to explore quantization and parameter-efficient fine-tuning approaches with larger-scale language models.

\paragraph{Realistic Web Setting.}
Our paper is partially motivated by the fact that LMs are trained on web content that increasingly contains synthetic text. However, after careful considerations, it is impossible to conduct experiments under a realistic web setting. To simulate a realistic setting, we would need a dataset of synthetic text posted on the web by real users. We initially thought about using data from ShareGPT where users upload their conversations with ChatGPT but then realized that this would pose copyright issues. It is thus not feasible to construct a realistic dataset for unconditional language modeling with synthetic content. In addition, there currently exists no algorithm that can reliably detect LM generated text and we cannot estimate the amount of synthetic information online. We have already proposed experiments with mixed settings, which demonstrated that the reduction in diversity only marginally lowered when mixing in a fixed percentage of human data. It would be interesting to conduct further experiments with varying combinations, potentially increasing the proportion of human data. Nevertheless, we anticipate that our research findings will continue to hold, given the minimal attenuation observed when experimenting with the current mixed setting.

\section*{Ethical Considerations}
\paragraph{Usage of scientific artifacts.}
We employ three datasets in our research: XL-SUM \citep{hasan-etal-2021-xl}, ACL Anthology\footnote{\url{https://aclanthology.org}}, and WritingPrompts \citep{fan-etal-2018-hierarchical}. XL-SUM and ACL Anthology are made available under the CC BY-NC-SA 4.0 license, while the WritingPrompts dataset is distributed under the MIT license. None of these datasets contains any information that can be linked to private individuals in a harmful way. Furthermore, we utilize the OPT model \citep{zhang2022opt}, which is subject to the ``OPT-175B License Agreement''\footnote{\url{https://github.com/facebookresearch/metaseq/blob/main/projects/OPT/MODEL_LICENSE.md}}. Our use of these resources aligns with their designated research purposes.

\paragraph{Potential risks.}
Our research focuses on the analysis of language models and is not specifically linked to any particular application. Its positive social impact lies in identifying and bringing to light overlooked issues in the usage of language models, thereby alerting both developers and users to exercise more deliberate considerations. However, it's important to recognize that there may be potential risks arising from the way our findings are interpreted by the general public, especially if they are exaggerated or overgeneralized. We want to stress that our conclusions are rigorously validated based on specific datasets and within a particular context. It is necessary to explicitly acknowledge these limitations when discussing our research during scientific dissemination.

\section*{Acknowledgements}
We thank Dr. Julie Hunter for her insights during the preliminary discussions of this work. We also thank the anonymous reviewers for their remarks and suggestions. This research has been partially supported by the ANR-TSIA HELAS chair and the ANR-23-CE23-0033-01 SINNet project.

\bibliography{anthology,custom}

\onecolumn
\appendix
\section{Visualization of Diversity Metrics}
\label{sec:appendix1}

\begin{figure}[h!]
    \centering
    \begin{subfigure}{.245\textwidth}
        \centering
        \includegraphics[width=\linewidth]{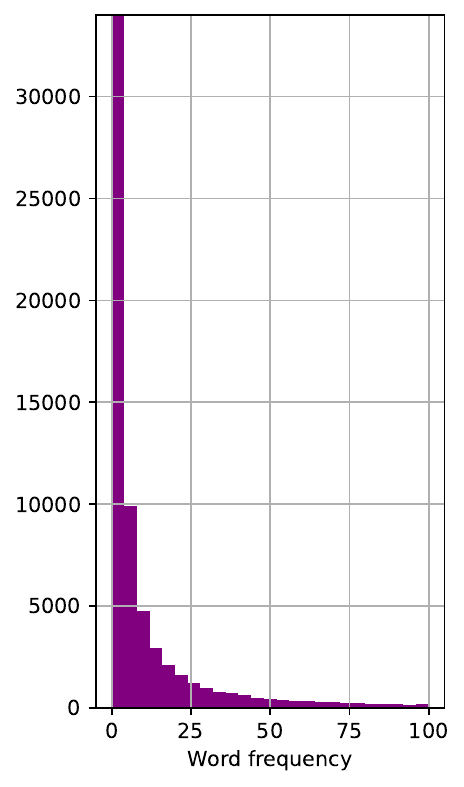}
        \caption{Human}
    \end{subfigure}%
    \hfill
    \begin{subfigure}{.245\textwidth}
        \centering
        \includegraphics[width=\linewidth]{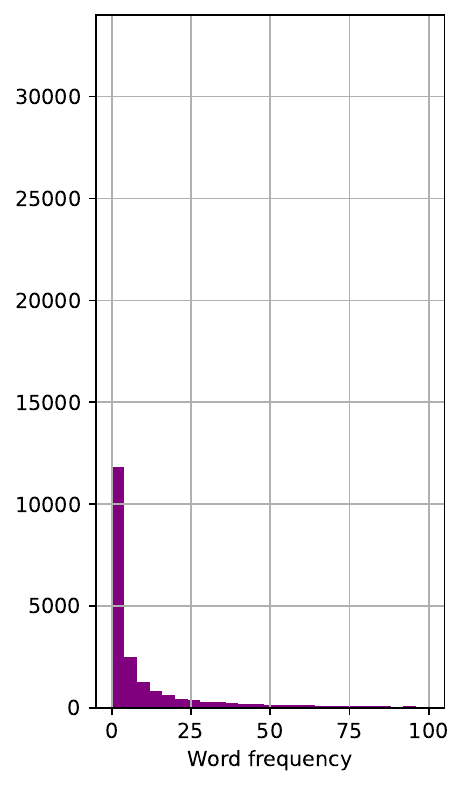}
        \caption{Iteration 1}
    \end{subfigure}
    \hfill
    \begin{subfigure}{.245\textwidth}
        \centering
        \includegraphics[width=\linewidth]{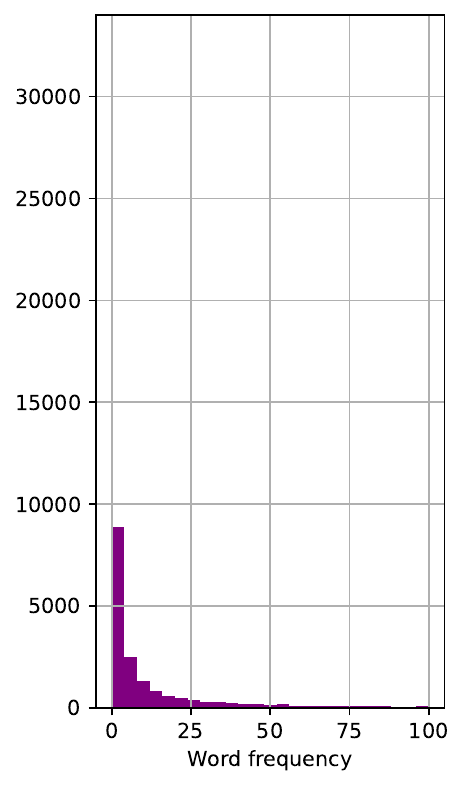}
        \caption{Iteration 3}
    \end{subfigure}
    \hfill
    \begin{subfigure}{.245\textwidth}
        \centering
        \includegraphics[width=\linewidth]{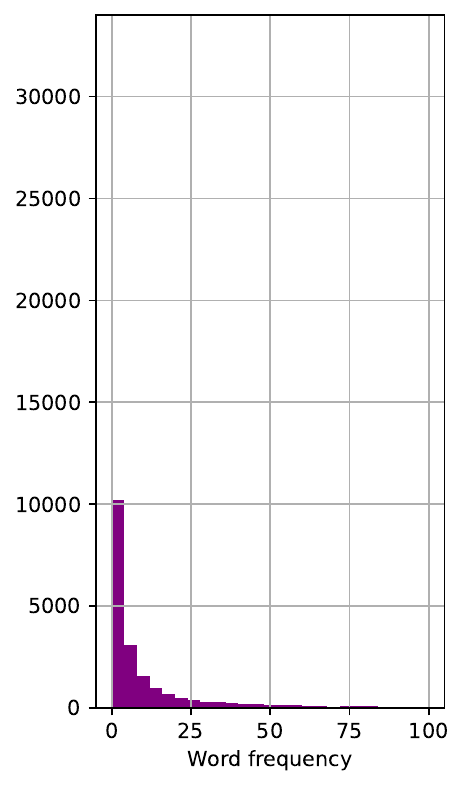}
        \caption{Iteration 5}
    \end{subfigure}

    \caption{Histograms illustrating word frequency in texts produced across various iterations for the story generation task. For visual clarity, the x-axis, representing word frequency, is truncated at 100, though the actual distribution extends further. A noticeable trend is the diminishing presence of low-frequency, ``unique'' words in the synthetic text relative to human-generated text, a pattern that intensifies with each iteration. This trend highlights a progressive decline of lexical diversity in the generated text.}
    \label{fig:tail}
\end{figure}

\vspace{1cm}

\begin{figure}[h!]
    \centering
    \includegraphics[width=0.87\textwidth]{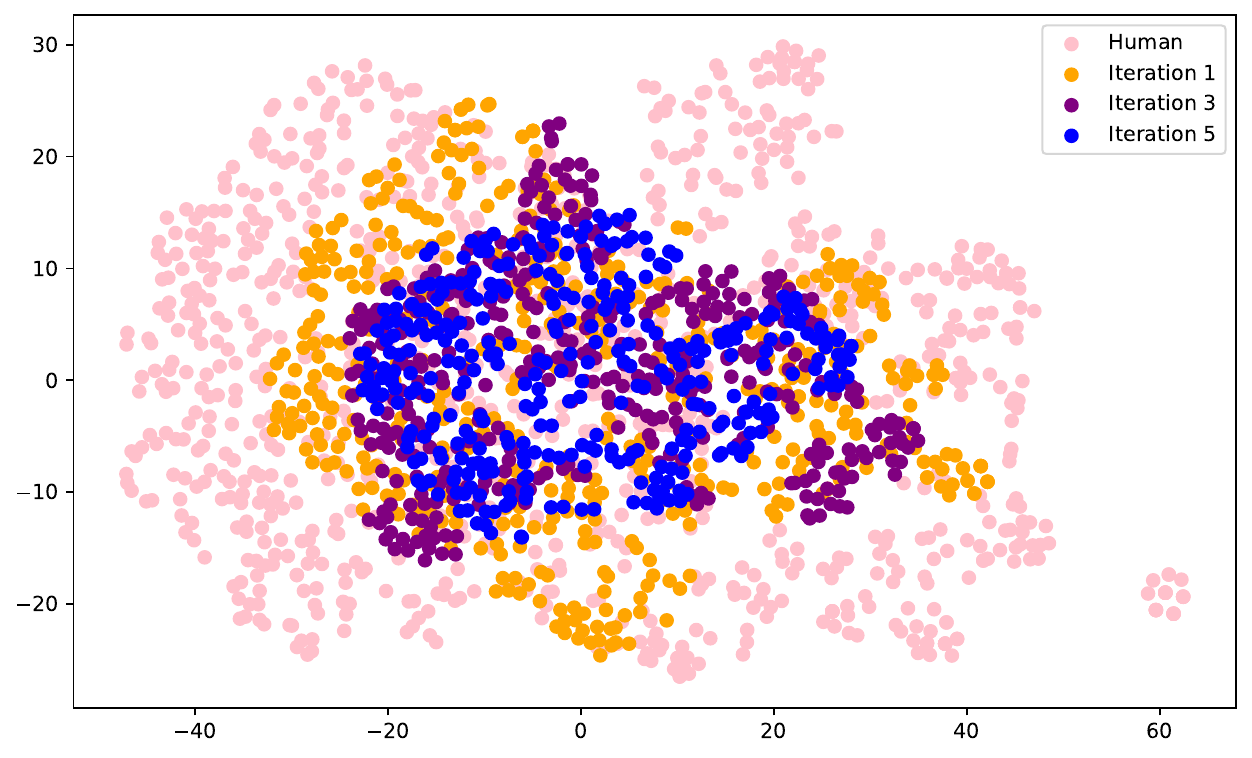}
    \caption{T-SNE visualization of dependency tree embeddings derived from sentences generated in successive iterations of our tuning-generation process. The visualization clearly depicts how, over time, the spatial distribution of the embeddings becomes increasingly compact. This decreasing spread is indicative of declining syntactic diversity. }
    \label{fig:syn}
\end{figure}

\vspace{1cm}

\begin{figure}[t!]
    \centering
    \includegraphics[width=0.87\textwidth]{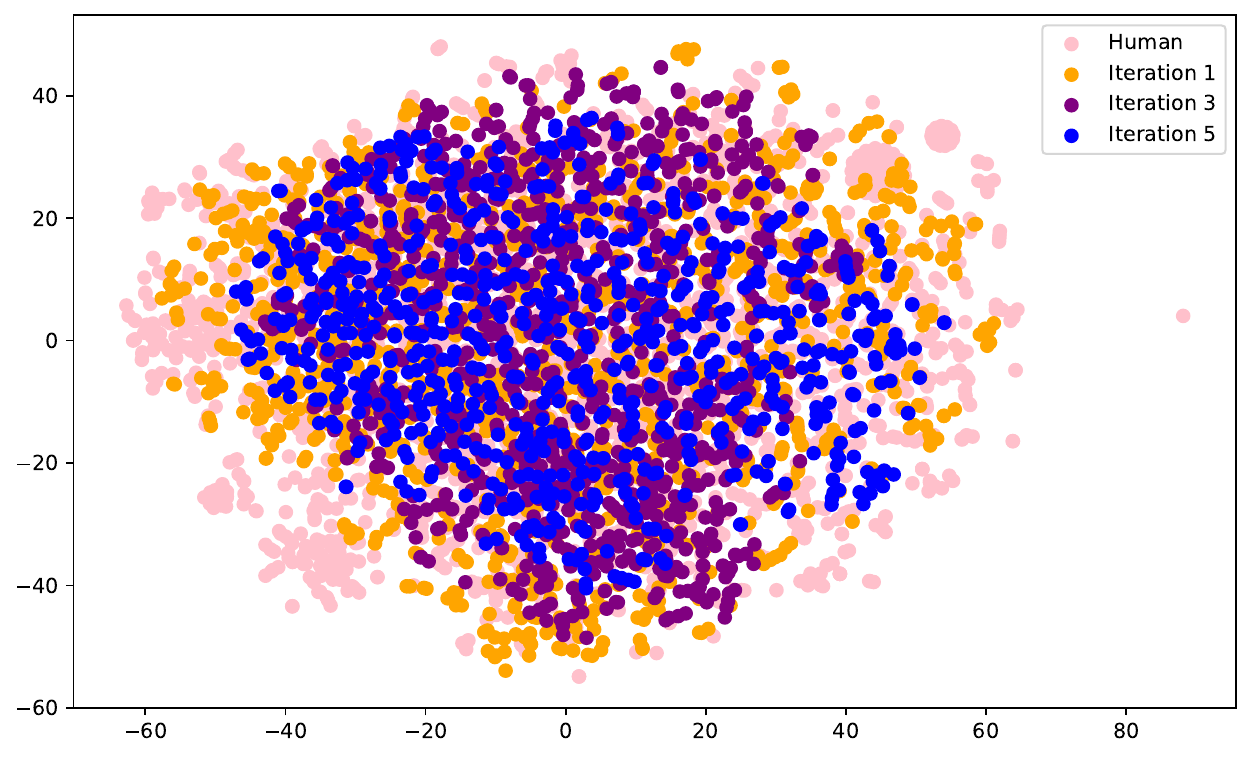}
    \caption{T-SNE visualization of sentence embeddings from text generated across different iterations. There is a noticeable decrease in dispersion over iterations, indicating a reduction in semantic diversity, though this change is less pronounced compared to that of syntactic diversity.}
    \label{fig:sem}
\end{figure}

\section{Implementation of Diversity Metrics}

\subsection{Preprocessing}
We apply preprocessing to the generated texts before computing the diversity metrics. For all three tasks, we remove the prompts from the generated texts. We remove <newline> tokens for story generation and replace URL links with WEBSITE for scientific abstract generation. We remove all punctuation marks for the calculation of lexical diversity metrics, but not for semantic diversity or syntactic diversity.

\subsection{Lexical Diversity Metrics}

\begin{table}[h]
\centering
\resizebox{\textwidth}{!}{%
\begin{tabular}{@{}llllllll@{}}
\toprule
\textbf{Iteration}             & {\color[HTML]{643403} \textbf{Human}} & 1      & 2      & 3      & 4      & 5      & 6      \\ \midrule
News Summarization             & 18.89                                 & 18.21  & 18.42  & 18.75  & 19.01  & 19.25  & 19.71  \\
Scientific Abstract Generation & 49.35                                 & 48.80  & 49.61  & 49.60  & 49.64  & 49.60  & 49.50  \\
Story Generation               & 148.77                                & 149.77 & 149.92 & 149.92 & 149.91 & 149.96 & 149.84 \\ \bottomrule
\end{tabular}%
}
\caption{Average text lengths for post-truncation generations. Text lengths are measured by the number of words.}
\label{tab:lengths}
\end{table}

The distinct-$n$ metric varies as a function of text length, so we compute results at fixed lengths. We apply truncation to the generated texts, using different thresholds for each task: 20 for news summarization, 50 for scientific abstract generation and 150 for story generation. The average text lengths after truncation are presented in Table \ref{tab:lengths}. We observe that all lengths consistently fall within a narrow range, allowing for fair comparison of distinct-$n$ results across iterations.

For Self-BLEU, we use a publicly available implementation\footnote{\url{https://github.com/Danial-Alh/fast-bleu}}. We take the mean value of Self-BLEU-2 and Self-BLEU-3.

\subsection{Semantic Diversity Metrics}
For sentence splitting, we use the NLTK sentence tokenizer. For Sentence-BERT, we use the all-mpnet-base-v2 model on huggingface. For the pairwise cosine distances, we randomly draw 2000 sentences for each calculation and report the mean value over 5 randomizations.

\subsection{Syntactic Diversity Metrics}
We construct the dependency graphs with the Stanza Dependency Parser\footnote{\url{https://stanfordnlp.github.io/stanza/depparse.html}}. We employ a publicly available implementation of the Weisfeiler-Lehman graph kernel\footnote{\url{https://github.com/ysig/GraKeL}}. We set the number of iterations to 2. The pairwise cosine distances are calculated in the same way as for semantic diversity.

\end{document}